\pdfoutput=1

\documentclass[11pt]{article}
\usepackage{acl}

\usepackage{times}
\usepackage{latexsym}
\usepackage{amsmath}
\usepackage{booktabs}
\usepackage{multirow}
\usepackage{pifont}
\usepackage{subcaption}
\usepackage{ascii}
\usepackage{mathtools}
\usepackage{array}
\usepackage{algorithm}
\usepackage{algpseudocode}
\usepackage{marvosym}
\usepackage[T1]{fontenc}
\usepackage[utf8]{inputenc}

\usepackage{microtype}
\DeclareMathOperator*{\argmax}{argmax}

\newcolumntype{M}[1]{>{\centering\arraybackslash}m{#1}}

\newcommand{\interalia}[1]{\citep[\emph{inter alia}]{#1}}
\newcommand{\Taskname}{Comparative opinion summarization}
\newcommand{\taskname}{comparative opinion summarization}
\newcommand{\corpus}{\textsc{CoCoTrip}}
\newcommand{\model}{\textsc{CoCoSum}}
\newcommand{\method}{Co-decoding}

\definecolor{c1}{HTML}{4e79a7}%
\definecolor{c2}{HTML}{f28e2b}%
\definecolor{c3}{HTML}{00AA55}%
\definecolor{c4}{HTML}{56B4E9}%
\definecolor{c5}{HTML}{CC79A7}%
\definecolor{c6}{HTML}{E69F00}%
\definecolor{c7}{HTML}{844E4D}%
\definecolor{c8}{HTML}{2D512A}%
\definecolor{c9}{HTML}{D81B60}
\definecolor{c10}{HTML}{1E88E5}
\definecolor{c11}{HTML}{400FA7}

\newcommand{\comm}[1]{{\color{c9}#1}}
\newcommand{\cont}[1]{{\color{c10} \textbf{#1}}}
\newcommand{\specific}[1]{{\color{c3} \textbf{#1}}}

\usepackage{tikz}

\title{Comparative Opinion Summarization via Collaborative Decoding}

\author{
Hayate Iso\textsuperscript{\Leo} \quad
Xiaolan Wang\textsuperscript{\Leo} \quad
Stefanos Angelidis\textsuperscript{\Jupiter} \quad
Yoshihiko Suhara\textsuperscript{\Leo} \\
\textsuperscript{\Leo}Megagon Labs
\textsuperscript{\Jupiter}University of Edinburgh \\
\texttt{\{hayate,xiaolan,yoshi\}@megagon.ai}
~~\texttt{s.angelidis@ed.ac.uk}
}

\begin{document}
\maketitle

\begin{abstract}
Opinion summarization focuses on generating summaries that reflect popular subjective information expressed in multiple online reviews.
While generated summaries offer general and concise information about a particular hotel or product, the information may be insufficient to help the user compare multiple different choices.
Thus, the user may still struggle with the question ``Which one should I pick?'' 
In this paper, we propose the {\em comparative opinion summarization} task, which aims at generating two contrastive summaries and one common summary from two different candidate sets of reviews.
We develop a comparative summarization framework \model{}, which consists of two base summarization models that jointly generate contrastive and common summaries.
Experimental results on a newly created benchmark \corpus{} show that \model{} can produce higher-quality contrastive and common summaries than state-of-the-art opinion summarization models.
The dataset and code are available at \url{https://github.com/megagonlabs/cocosum}.
\end{abstract}

\section{Introduction}
\begin{figure}[t]
    \centering
    \includegraphics[width=0.99\linewidth]{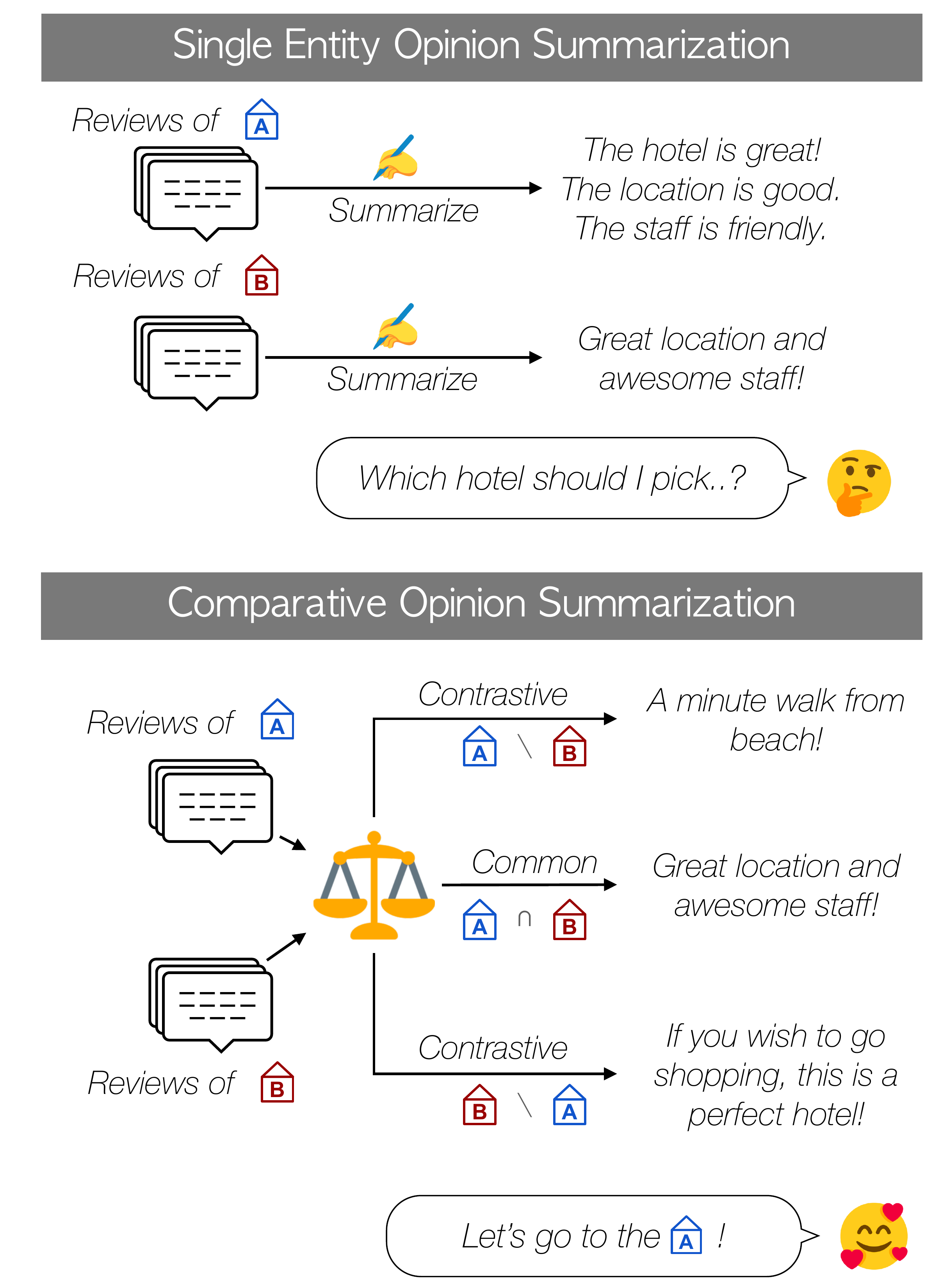}
     \caption{Overview of the \taskname{} task. The model takes two set of reviews about different entities to generate two contrastive opinion summaries, which contain  distinctive opinions, and one common opinion summary, which describes common opinions between the two entities.}
    \label{fig:task}
\end{figure}
\begin{figure*}
    \centering
    \includegraphics[width=0.99\textwidth]{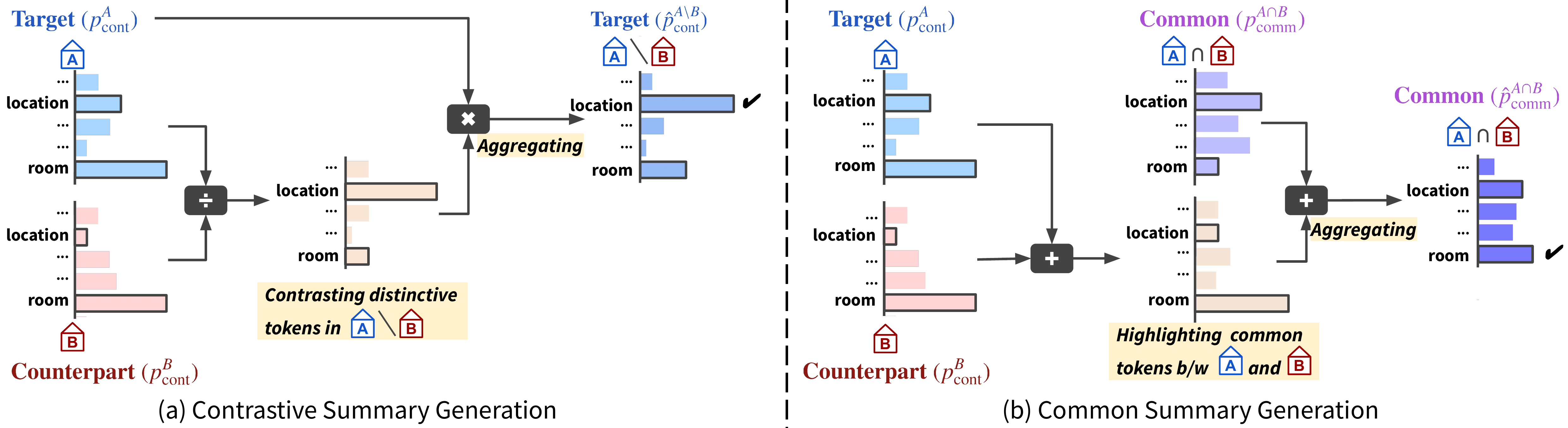}
    \caption{%
    Illustration of \method{}: (a) For contrastive summary generation, distinctive words are emphasized by {\em contrasting} the token probability distribution of target entity against that of the counterpart entity. (b) For common summary generation, entity-pair-specific words are highlighted by {\em aggregating} token probability distributions of all base models to alleviate the overly generic summary generation issue.
    }
    \label{fig:codec}
\end{figure*}

Widely available online customer reviews help users with decision-making in a variety of domains (e.g., hotel, restaurant, or company). After creating a list of candidate choices based on initial conditions (e.g., area, price range, restaurant type), the user often has to compare a few choices in depth by carefully reading the reviews to make a final decision. However, it is time-consuming and difficult for the user to detect differences and similarities between the candidates, as those pieces of information are often scattered in different reviews.

The recent success of neural summarization techniques and the growth of online review platforms led to establishing the field of multi-document opinion summarization~\cite{chu2019meansum,brazinskas-etal-2020-unsupervised,amplayo-lapata-2020-unsupervised,suhara-etal-2020-opiniondigest,iso21emnlpfindings}. The goal of multi-document opinion summarization is to generate a summary that represents salient opinions in input reviews of a particular hotel or product, which we refer to as an entity.
However, existing opinion summarization techniques are designed to generate a {\it single-entity opinion} summary that reflects popular opinions for each entity, without taking into account {\it contrastive and common opinions} that are uniquely (commonly) mentioned in each entity (both entities) as depicted in Figure~\ref{fig:task}.
Therefore, the user still needs to figure out which opinions are contrastive or common between the entities by carefully reading and comparing summaries generated by existing opinion summarization solutions.

To this end, we take one step beyond the current scope of opinion summarization and propose a novel task of generating contrastive and common summaries by comparing multiple entities, which we refer to as {\em \taskname}. 
In contrast to the conventional single-entity opinion summarization task that makes a general summary for each entity, the goal of \taskname{} is to generate two contrastive summaries and one common summary from two sets of reviews about two entities. Thus, the user can easily understand distinctive and common opinions about multiple entities. In this paper, we consider pairwise comparison as it is the most common choice and the minimal unit for multiple comparisons.

A key challenge of building a summarizer for the task is that the model has to correctly distinguish what contrastive and common opinions from input reviews of two entities are. 
Existing opinion summarization models do not implement this functionality as they are designed to summarize popular opinions for a single entity.

To address this issue, we develop a \taskname{} framework \model{}, which consists of two base summarization models for contrastive and common opinion summary generation.
\model{} employs a novel Collaborative Decoding (\method) algorithm that takes two review sets as input to \textit{compare and contrast} the token probability distributions of the models to generate more distinctive summaries as illustrated in Figure~\ref{fig:codec}.

Experimental results on a newly created comparative opinion summarization benchmark \corpus{} show that \model{} with \method{} generate substantially high-quality contrastive and common summaries compared to baseline models including state-of-the-art opinion summarization models.

Our contributions are as follows:
\begin{itemize}
  \setlength{\parskip}{0cm}
  \setlength{\itemsep}{0cm}
  \item We propose the novel task of \taskname, which takes two review sets as input and outputs two contrastive summaries and one common summary.
  \item We develop \model{}, which consists of two base summarization models and implements a novel \method{} algorithm that facilitates generating distinctive and entity-pair-specific summaries by aggregating the token probability distributions of the models.
  \item We create and release a \taskname{} benchmark \corpus{} that contains manually written reference summaries for 48 entity pairs.
\end{itemize}

\section{Comparative Opinion Summarization}
\subsection{Problem Formulation}
Let $\mathcal{C}$ be a corpus of reviews on entities from a single domain (e.g., hotels). For each entity $e$, we define its review set $\mathcal{R}_e = \{r_{e,1}, r_{e, 2}, \dots, r_{e, |\mathcal{R}_e|}\}$.

We define {\em contrastive opinions} of a target entity $A$ against a counterpart entity $B$ as subjective information that is described only in $\mathcal{R}_A$ but not in $\mathcal{R}_B$ and refer to the summary that contains such contrastive opinions as a {\em contrastive summary} $y_{\text{cont}}^{A\setminus B}$.
Similarly, we define {\em common opinions} of entities $A$ and $B$ as subjective information that is commonly described in $\mathcal{R}_A$ and $\mathcal{R}_B$ and refer to the summary that contains common opinions as a {\em common summary} $y_{\text{comm}}^{A\cap B}$.
Note that $y_{\text{comm}}^{A\cap B}$ and $y_{\text{comm}}^{B\cap A}$ are identical, thus we consider a single common summary for an entity pair.

We formalize {\em \taskname{}} as a task to generate two sets of contrastive summaries $y_{\text{cont}}^{A\setminus B}$,  $y_{\text{cont}}^{B\setminus A}$, and one common summary $y_{\text{comm}}^{A\cap B}$ from two sets of reviews $\mathcal{R}_{A}$ and $\mathcal{R}_{B}$ for a pair of entities $A$ and $B$. Compared to existing summarization tasks, comparative opinion summarization is the first work that aims to generate abstractive summaries for contrastive and common opinions.

\begin{table*}[t]
    \centering\footnotesize
    \begin{tabular}{l|cccc|cccc}\toprule
        & \multirow{2}{*}{Task} & Input & Summary & \multirow{2}{*}{Domain}& \multicolumn{4}{c}{\% of novel $n$-grams in gold summary} \\
        & & length & length & & unigram & bigram & trigram & 4-gram \\\midrule
        \multirow{2}{*}{\corpus{}~(Ours)} & Contrastive & \multirow{2}{*}{1529.4} & 132.9 & \multirow{2}{*}{Hotels} & 22.81 & 72.41 & 91.43 & 97.08\\
        & Common & & 20.3 & & 9.27 & 51.75 & 84.52 & 95.75\\\midrule
        \citet{chu2019meansum} & Single & 581.1 & 70.4 & Businesses & 30.87 & 83.23 & 96.60 & 99.18\\
        \citet{brazinskas-etal-2020-unsupervised} & Single & 473.4 & 59.8 & Products & 26.23 & 77.52 & 93.24 & 97.43\\
        \citet{angelidis-et-al-2021-qt} & Single & 16160.6 & 83.6 & Hotels & 1.98 & 21.13 & 47.14 & 63.86\\
        \bottomrule
    \end{tabular}
    \caption{Statistics of \corpus{} and other benchmarks. \corpus{} has a comparable corpus size against the benchmarks while offering unique characteristics (i.e., three types of reference summaries for a pair of entities). The average input length in tokens 
    is calculated using concatenated input reviews.}
    \label{tab:dataset_stats}
\end{table*}

\subsection{The \corpus{} Corpus}
As the task requires three types of reference summaries for each {\em entity pair}, none of the existing benchmarks for single-entity opinion summarization can be used for evaluation.
Therefore, we create a \taskname{} corpus \corpus{} that contains human-written contrastive and common summaries for 48 pairs of entities. We sampled the entity pairs and reviews from the TripAdvisor corpus~\cite{wang-et-al-2010-tripadvisor-corpus}.

We sampled 16 reviews for every pair (i.e., 8 reviews for each entity). For every entity pair, we collected 3 gold-standard summaries written by different annotators for two contrastive summaries and one common summary.
Details of the corpus creation process are described in Appendix.

We summarize the \corpus{} dataset and compare it with existing opinion summarization datasets in Table~\ref{tab:dataset_stats}.
We calculate novel $n$-grams in gold summaries to evaluate how abstractive/extractive \corpus{} is. Considering the input and summary length, we confirm that \corpus{} is sufficiently abstractive compared to the existing opinion summarization datasets.

\section{\model}
In order to summarize contrastive and common opinions from two sets of reviews, the comparative opinion summarization task requires the model to compare and contrast two sets of reviews; however, existing single-entity opinion summarization models do not have such functionality.
Therefore, we design a ``collaborative'' decoding solution \method, which characterizes the target summary distribution by leveraging two base summarization models.

\subsection{Base Summarization Model}
\label{sec:design}

\begin{figure}
    \centering
    \includegraphics[width=0.99\linewidth]{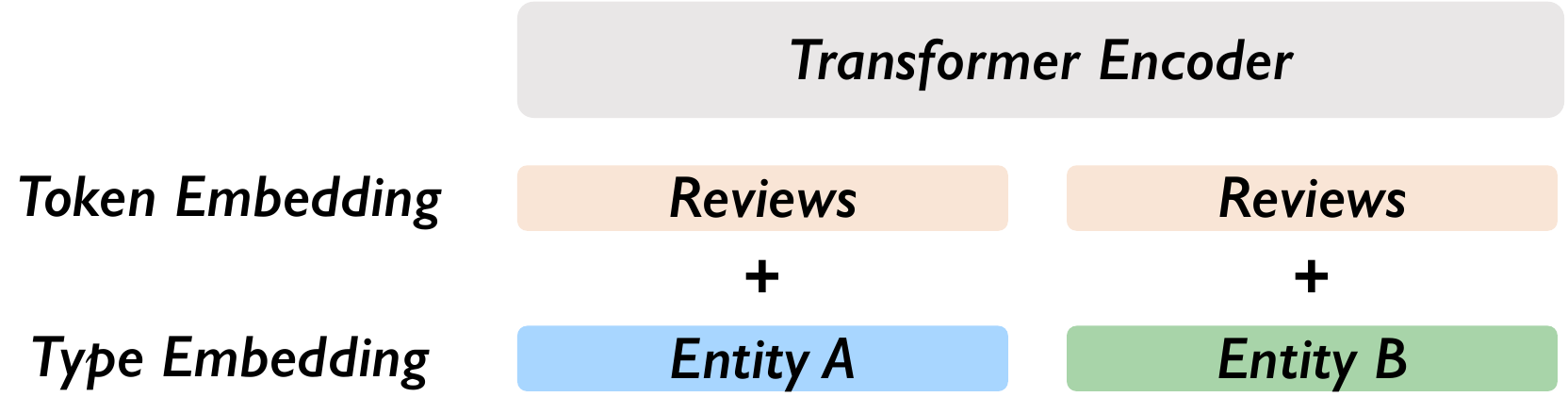}
    \caption{
    Encoder of the
    base
    common
    summarization model has {\em type embeddings} to distinguish the original entity.%
    }
    \label{fig:encoding}
\end{figure}

\model{} consists of two base summarization models.
The base contrastive summarization model is a single-entity summarization model that takes only reviews of the target entity as input, while the base common summarization model takes reviews of two entities as input.
In both cases, the input reviews are concatenated into a single sequence before encoding.
To help the encoder distinguish the entity, we introduce additional {\em type embeddings} into the input layer of the encoder as shown in Figure~\ref{fig:encoding}.

For common summary generation (i.e., $y_{\text{comm}}^{A\cap B} = y_{\text{comm}}^{B\cap A}$), the model should generate the same common summary for the same entity pair regardless of the input order of review sets. 
Thus, we create two input sequences (i.e., $A \cap B$ and $B \cap A$) and merge the token probability distributions of the two sequences for a summary generation.

\subsection{Collaborative Decoding}%
\label{sec:codecoding}
As illustrated in Figure~\ref{fig:codec}, Co-decoding combines predictions of the target and the counterpart (and common, for common summary generation) opinion summarization models during the inference time.
The key idea of \method{} is to aggregate token probability distributions of contrastive summarization model $p_{\text{cont}}(\cdot)$ and common summarization model $p_{\text{comm}}(\cdot)$ at each step, so the two models can collaboratively generate (1) contrastive summaries that contain distinctive opinions that do not appear in the counterpart review set and (2) common summaries that only contain common opinions that appear in both target and counterpart review sets.

\noindent
\paragraph{Contrastive summary generation}
To improve the distinctiveness of generated contrastive summaries that only contains entity-specific opinions, we consider {\em penalizing} the tokens that are likely to appear in the counterpart entity. That is, we use two token probability distributions and highlight tokens that are distinctive compared to the counterpart entity by using the {\em token ratio distribution} between them.
We also introduce a trade-off hyperparameter $\delta$ that controls the balance between the original token distribution and the token ratio distribution:
\begin{align}\label{eq:codec_cont}
    \hat{p}_{\text{cont}}^{A\setminus B}(y_t) \propto p_{\text{cont}}^{A}(y_t)\left(\frac{p_{\text{cont}}^{A}(y_t)}{p_{\text{cont}}^{B}(y_t)}\right)^{\delta},
\end{align}
where we denote the token probability distribution of the base contrastive summarization model given the previously generated tokens $y_{<t}$ and the set of input reviews $\mathcal{R}_e$ for entity $e \in \{A, B\}$ at $t$-th step by $p_{\text{cont}}^{e}(y_t):=p_{\text{cont}}(y_t \mid y_{<t}, \mathcal{R}_e)$.
Note that for both $p_{\text{cont}}^{A}(y_t)$ and $p_{\text{cont}}^{B}(y_t)$, we use the same prefix $y_{<t}$. For the other contrastive summary $\hat{y}_{\text{cont}}^{B \setminus A}$, the token probability can be obtained by
swapping $A$ and $B$ in Eq. (\ref{eq:codec_cont}).

\method{} for contrastive summary generation is illustrated in Figure~\ref{fig:codec} (a).
The intuition behind this approach is that the token ratio distribution $\frac{p_{\text{cont}}^{A}(y_t)}{p_{\text{cont}}^{B}(y_t)}$ (i.e., $A \land \neg B$) highlights distinctive tokens that are relatively unique to the target entity, which are emphasized by combining with the original token distribution.  
This can be considered a variant of Product-of-Experts (PoE)~\cite{hinton2002training,liu-etal-2021-dexperts}, which models Logical AND with multiple probabilistic distributions.

\noindent
\paragraph{Common summary generation}
Common summaries should contain common opinions that are about a given pair of entities.
However, we observe that simply fine-tuned summarization models tend to generate overly generic summaries that can be true for any entity pair.

To incorporate the entity-specific information into the common summary,
we design \method{} to use the sum of the token probability distributions of the contrastive summarization model, which is then combined with the original token probability distribution using a trade-off hyperparameter $\gamma$:

\begin{align}\label{eq:codec_comm}
    \hat{p}^{A\cap B}_{\text{comm}}(y_t) \propto p^{A\cap B}_{\text{comm}}(y_t) + \gamma \smashoperator{\sum_{E \in \{A, B\}}} p_{\text{cont}}^{E}(y_t),
\end{align}
where we denote the token probability distribution of the base common summarization model by $p_{\text{comm}}^{A\cap B}(y_t) :=p_{\text{comm}}(y_t \mid y_{<t}, \mathcal{R}_A,  \mathcal{R}_B)$.

\method{} for common summary generation is illustrated in Figure~\ref{fig:codec} (b).
The intuition behind this approach is that we first identify salient tokens for the input entity pair by adding the token probability distributions of contrastive summaries: $p_{\text{cont}}^{A}(y_t) + p_{\text{cont}}^{B}(y_t)$ (i.e., $A \lor B$), which is then combined with the original distribution using the trade-off hyperparameter $\gamma$. 
This can be considered a variant of Mixture-of-Experts (MoE)~\cite{Jacobs1991AdaptiveMO}, which models Logical OR with multiple probabilistic distributions and is suitable for {\em interpolating} the token probability distribution of models with different characteristics.

We would like to emphasize that \method{} is a token probability distribution calculation method for \taskname{} based on two summarization models; thus, it is flexible of the choice of the base summarization model and the decoding algorithm.

\begin{table*}[t]
    \centering
    \small
    \addtolength{\tabcolsep}{-1pt} 
    \begin{tabular}{l|cccc|cccc|c}
        \toprule
        & \multicolumn{4}{c}{\textbf{Contrastive}} & \multicolumn{4}{c|}{\textbf{Common}} & \textbf{Pair}\\
         & R1 & R2 & RL & BS & R1 & R2 & RL & BS & DS \\\midrule
        \textbf{Self-supervised}\\
        \ \textit{Extaractive models}\\
        \quad LexRank$_{\textsc{TFIDF}}$~\tiny{\cite{erkan2004lexrank}} & 35.38 & 7.39 & 18.25 & 22.61 & 22.51 & 4.00 & 15.26 & 24.65 & 63.28\\
        \quad LexRank$_\textsc{BERT}$~\tiny{\cite{reimers-gurevych-2019-sentence}} & 32.65 & 5.67 & 16.67 & 20.51 & 17.91 & 2.95 & 12.60 & 24.83 & 65.56\\
        \ \textit{Abstractive models}\\
        \quad MeanSum~\tiny{\cite{chu2019meansum}} & 34.19 & 7.84 & 19.76 & 23.89 & 13.09 & 0.85 & 10.41 & 16.13 & 65.98 \\
        \quad OpinionDigest~\tiny{\cite{suhara-etal-2020-opiniondigest}} & 37.30 & 8.67 & 20.36 & 21.77 & 21.52 & 4.41 & 15.26 & 17.06 & 64.87\\
        \quad CopyCat~\tiny{\cite{brazinskas-etal-2020-unsupervised}} & 35.30 & 8.39 & 18.64 & 21.91 & 36.16 & 11.91 & 25.15 & 50.16 & 40.80\\ %
        \quad BiMeanVAE~\tiny{\cite{iso21emnlpfindings}} & 37.44 & 9.41 & 22.02 & 24.33 & 38.47 & 14.17 & 27.46 & 50.98 & 42.55\\\midrule
        \textbf{CoCoSum} (Ours) \\
        \quad Self-supervised & 40.01 & \textbf{10.80} & \textbf{21.97} & \textbf{30.02} & \textbf{41.13} & \textbf{15.25} & \textbf{30.60} & \textbf{54.65} & \textbf{66.00}\\ %
        \qquad w/o \method{} \tiny{($\delta = \gamma = 0.$)} & \textbf{40.78} & 10.66 & 21.53 & 29.90 & 40.40 & 14.13 & 29.81 & 54.28 & 57.63\\
        \quad Few-shot & 42.22 & 12.11 & 24.13 & \textbf{35.63} & \textbf{46.80} & \textbf{20.68} & \textbf{35.62} & \textbf{61.52} & \textbf{74.02} \\ %
        \qquad w/o \method{} \tiny{($\delta = \gamma = 0.$)} & \textbf{43.65} & \textbf{12.83} & \textbf{24.93} & 35.42 & 45.90 & 19.59 & 34.40 & 59.32 & 71.69\\
        \midrule %
        Human upper bound & 47.37 & 13.00 & 26.03 & 37.69 & 52.26 & 19.16 & 39.89 & 61.10 & 71.79\\
         \bottomrule
    \end{tabular}
    \caption{ROUGE and BERT scores (summarization quality) for contrastive and common summaries on \corpus{} and the distinctiveness score (DS) of generated summaries.
    CoCoSum significantly improves the distinctiveness while keeping high summarization quality.
    Human upper bound is measured by calculating the corresponding score across multiple reference summaries.}
    \label{tab:main}
\end{table*}

\section{Evaluation}
\subsection{Experimental Settings}
We build two versions of \model{} using self-supervised training (Self-supervised) and few-shot learning (Few-shot). We evaluate the summarization performance of the two versions with and without \method{}.
For all the base models, we use a pre-trained LED model~\cite{Beltagy2020Longformer}, which uses sparse attention to handle long sequences and thus is suitable for the purpose.\footnote{\url{https://huggingface.co/allenai/led-base-16384}}

For self-supervised training, we use the TripAdvisor review corpus~\cite{wang-et-al-2010-tripadvisor-corpus} to construct pseudo review-summary pairs following \citet{elsahar-etal-2021-self} with two modifications:
1) We filter reviews with different word length ranges for contrastive ($[100, 150]$) and common ($[15, 50]$) base models to accommodate the different average summary lengths.
2) For the self-supervised base common summarization model, as it takes two sets of reviews (i.e., $\mathcal{R}_A$,  $\mathcal{R}_B$) as input, we retrieve and merge review-summary pairs, based on the summary similarity, to make a pseudo training dataset. 

For few-shot learning, we use 20 instances of \corpus{} for further fine-tuning self-supervised base summarization models. Detailed analysis of the few-shot learning strategies can be found in Appendix. %

For evaluation, we used the remaining 10 instances of \corpus{} for development and 18 instances for testing.

For \method, we used top-$p$ vocabulary~\cite{Holtzman2020The}, which is the smallest token set whose cumulative probability exceeds $p$, with $p=0.9$ for $p_{\text{cont}}^{A}(y_t)$, $p_{\text{cont}}^{B}(y_t)$, and $p_{\text{comm}}^{A \cap B}(y_t)$.
We used Beam Search with a width of 4. We chose $\delta$ and $\gamma$ using the dev set.

\begin{table*}[ht]
    \centering
    \small
    \begin{tabular}{l|ccc|ccc|ccc}
        \toprule
        & \multicolumn{3}{c}{Content overlap} & \multicolumn{3}{|c}{Content support} & \multicolumn{3}{|c}{Quality}\\
        & Overlap $\downarrow$ & Partial $\downarrow$ & Not $\uparrow$& Full $\uparrow$& Partial $\uparrow$ & No $\downarrow$ & Coh $\uparrow$& Info $\uparrow$& Non-red $\uparrow$\\\midrule
        BiMeanVAE
        & 64.45 & 20.19 & 15.35 & 45.23 & 31.54 & 23.24 & 3.78 & 2.34 & 3.11\\
        OpinionDigest
        & 20.73 & 21.15 & 58.12 & 42.31 & 28.53 & 29.17 & 3.53 & 2.28 & 3.29\\
        \model$_\text{few}$ & \textbf{4.80} & \textbf{25.20} & \textbf{70.00} & \textbf{63.50} &  \textbf{24.09} & \textbf{12.41} & 4.10 & \textbf{2.81} & \textbf{4.38}\\
        \quad w/o \method & 10.02 & 22.14 & 67.84 & 58.27& 25.90 & 15.83& \textbf{4.19} & 2.80 & 4.34\\
    \bottomrule
    \end{tabular}
    \caption{Human evaluations on content overlap, content support, coherence (coh.), informativeness (info.), and non-redundancy (non-red).}
    \label{tab:human_eval}
\end{table*}

To access the quality of \model{}, we evaluated the performance of a variety of baseline approaches:\\ \textbf{LexRank}$_{\text{TFIDF}}$~\cite{erkan2004lexrank}: The classic unsupervised opinion summarization solution; \\
\textbf{LexRank}$_{\text{BERT}}$~\cite{erkan2004lexrank,reimers-gurevych-2019-sentence}: LexRank approach with Sentence BERT~\cite{reimers-gurevych-2019-sentence} embeddings\footnote{\url{https://github.com/UKPLab/sentence-transformers}};\\
\textbf{MeanSum}~\cite{chu2019meansum}: the unsupervised single entity opinion summarization solution\footnote{\url{https://github.com/sosuperic/MeanSum}}; \\
\textbf{CopyCat}~\cite{brazinskas-etal-2020-unsupervised}: a single entity opinion summarization solution based on leave-one-out reconstruction\footnote{\url{https://github.com/abrazinskas/Copycat-abstractive-opinion-summarizer}}; \\ \textbf{BiMeanVAE}~\cite{iso21emnlpfindings}: an optimized single entity opinion summarization solution\footnote{\url{https://github.com/megagonlabs/coop}} for MeanSum.

For those baseline models above, we use $\mathcal{R}_{A}$ (or $\mathcal{R}_{B}$) as input for the contrastive summary and both $\mathcal{R}_{A}$ and $\mathcal{R}_{B}$ as input for the common summary.\\

\smallskip

\noindent \textbf{OpinionDigest}~\cite{suhara-etal-2020-opiniondigest}: a weakly supervised opinion summarization approach.\footnote{\url{https://github.com/megagonlabs/opiniondigest}} 
We customize OpinionDigest for comparative opinion summarization. Specifically, we categorize opinion clusters extracted from $\mathcal{R}_{A}$ and $\mathcal{R}_{B}$ as ``contrastive'' if the cluster only contains opinions from a single entity and ``common'' if the cluster contains opinions from both of the entities. In this way, OpinionDigest can leverage the extracted opinion clusters to generate contrastive and common summaries. %
\\

\subsection{Automatic Evaluation}
\noindent
\paragraph{Evaluation metrics}
For summarization quality, we use ROUGE 1/2/L F1 scores~\cite{lin-2004-rouge}\footnote{ \url{https://github.com/Diego999/py-rouge}} and BERTScore~\cite{bert-score}\footnote{DeBERTa NLI model~\cite{he2021deberta} and baseline re-scaling are used.} as automatic evaluation based on reference summaries.

To evaluate the {\em distinctiveness} of generated summaries, we calculate the average distinctiveness score (DS) between generated contrastive summaries and common summaries for all entity pairs defined as follows:
\begin{align*}
    \text{DS} = 1 - \frac{\sum_{(y, z) \in \hat{\mathcal{Y}}^{(2)}}|\mathcal{W}_y \cap \mathcal{W}_z| - 2 |\bigcap_{y \in \hat{\mathcal{Y}}} \mathcal{W}_{y}|}{|\bigcup_{y \in \hat{\mathcal{Y}}} \mathcal{W}_{y}|},
\end{align*}
where $\hat{\mathcal{Y}} := \{\hat{y}_{\text{cont}}^{A\setminus B}, \hat{y}_{\text{cont}}^{B\setminus A}, \hat{y}_{\text{comm}}^{A\cap B}\}$, $\mathcal{W}_{y}$ is the token bag of generated summary $y \in \hat{\mathcal{Y}}$, and $\hat{\mathcal{Y}}^{(2)}$ is the 2-subsets of $\hat{\mathcal{Y}}$. The DS will be higher if the word overlaps between two generated contrastive summaries $\hat{y}_{\text{cont}}^{A\setminus B}$, $\hat{y}_{\text{cont}}^{B\setminus A}$, and a generated common summary $\hat{y}_{\text{comm}}^{A\cap B}$ are smaller.

\smallskip

\paragraph{Results}
As shown in Table~\ref{tab:main}, \model{} outperforms the baseline methods for the ROUGE and BERT scores (summarization quality) and the distinctiveness score (DS), showing the effectiveness of our self-supervised dataset and \method.
Comparing the summarization quality by \model{} and \model{} w/o \method, we confirm that \method{} significantly improves the distinctiveness especially in self-supervised setting while maintaining the summarization performance.

Among the baseline methods, BiMeanVAE shows the highest ROUGE scores while performing poorly for the distinctiveness score. 
Although MeanSum and OpinionDigest show high distinctiveness scores, they show significantly worse performance on the common summary generation task. 
The results indicate it is challenging for existing opinion summarization models to improve the distinctiveness of generated summaries while keeping them high-quality for both of the tasks. 

\subsection{Human Evaluation}
For human evaluation, we hired contractors from Upwork\footnote{\url{https://www.upwork.com}} platform and conducted three sets of human evaluation comparing \model{} with two representative baselines---BiMeanVAE and OpinionDigest. 

First, we asked the human annotators to evaluate the overlapped content between the contrastive summaries and the common summary for a given entity pair. More specifically, for every sentence in the summary, we asked human annotators to judge if its content is {\em overlap}, {\em partially overlap}, or {\em not overlap} with the other two summaries. According to the problem formulation, less overlap, i.e., not or partially overlap, is preferred. As shown in Table~\ref{tab:human_eval}, \model{} is significantly better than \model{} w/o \method, and is substantially better than BiMeanVAE and OpinionDigest. This result also aligns with our automatic evaluation on the distinctiveness (DS in Table~\ref{tab:main}), and it demonstrates that \model{} can produce more {\em distinctive} contrastive and common summaries.

Second, we conducted a summary content support study to evaluate how faithful the generated summaries are toward the input reviews. Similar to content overlap, for every sentence in the summary, we asked human annotators to judge if its content is {\em fully supported}, {\em partially supported}, or {\em not supported} by the corresponding input review sentences.
Note that the input review sentences are selected based on sentence-level labels we acquired from \corpus{}. The results show that \model{} is able to generate the most faithful summaries compared to all the other baselines. 

Lastly, we asked the human annotators to give ratings (from 1 to 5) for the generated summaries with respect to three criteria, namely {\em coherence}, {\em informativeness}, and {\em non-redundancy}. We report the average ratings~\cite{harpe2015analyze} for the summaries generated from different methods in Table~\ref{tab:human_eval}. As shown in the table, summaries generated by \model{} is slightly less coherent than \model{} w/o \method. This slight degradation is expected because \method adjusts the token probability to encourage contrastive/common content, thus it may also prioritize tokens that are less coherent. Other than coherence, \model{} shows slightly better informativeness and non-redundancy. Meanwhile, compared to BiMeanVAE and OpinionDigest, \model{} shows much better performance on all the three criteria.

\section{Analysis}\label{sec:analysis}
\subsection{Distinctiveness in Generated Summaries}
In addition to the summarization quality, distinctiveness is another important factor for \taskname{} to help the user pick one against the other. Therefore, we conduct additional analysis to investigate the quality of distinctiveness in generated summaries. 

\noindent
\paragraph{How distinctive are generated contrastive summaries for each entity pair?}

To complement our experiments on the distinctiveness score (in Table~\ref{tab:main}), which considers both types of generated summaries, we further evaluate {\em intra-entity-pair BERTScore (Intra-BERTScore)} only between two contrastive summaries for each entity pair to measure the {\em intra-entity-pair distinctiveness} defined by the average of $\text{BERTScore}(\hat{y}_{\text{cont}}^{A\setminus B}, \hat{y}_{\text{cont}}^{B\setminus A}) $.

\begin{figure}[t]
    \centering
    \includegraphics[width=\linewidth]{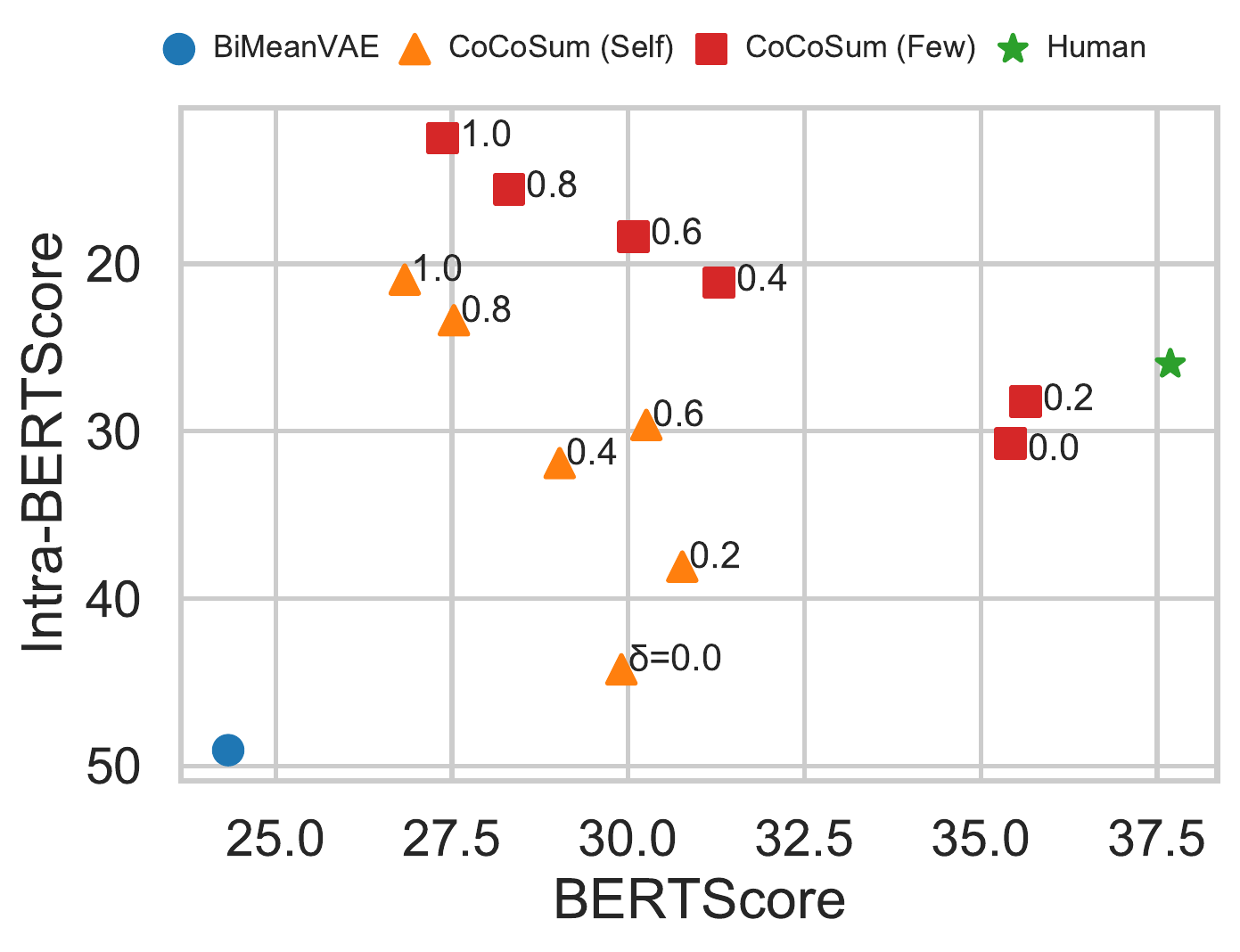}
    \caption{BERTScore and Intra-BERTScore for generated contrastive summaries with different hyperparameters $\delta$. The goal is to generate high quality and distinctive summaries (upper right).}
    \label{fig:intra_bs}
\end{figure}

Figure~\ref{fig:intra_bs} shows that in both self-supervised and few-shot settings, \model{} significantly outperforms the state-of-the-art opinion summarization model (BiMeanVAE).
The results confirm that \method{} successfully generates more distinctive opinions of each other, and the hyperparameter $\delta$ controls the trade-off between the summarization quality (BERTScore) and the distinctiveness (Intra-BERTScore).

\begin{table*}[t]
    \centering
    \small
    \begin{tabular}{p{7cm}|p{7cm}}
        \toprule
        \multicolumn{2}{l}{\textbf{\model}}\hfill Intra-ROUGE1/2/L = (\textbf{36.84, 8.51, 23.16})\\\midrule
        \textbf{Entity ID: 305813} & \textbf{Entity ID: 305947} \\
        ~\cont{The Langham Place Hotel is a 4-star hotel.} \comm{It is centrally located} \cont{and has easy access to the mall and cinema next door. The room was lovely with a great view. The bed in the room was firm and not too comfy. The spa facilities here at the hotel are of a really high standard.} \comm{The staff at this hotel are generally excellent and very much co-operative.} \cont{The hotel had over priced buffet meals and snacks and drinks but the club floor facilities are of such a high standard that you know you are worth it.} & ~\cont{The Metropark Kowloon is a good hotel to stay at for a week or longer. It's ideally located for those who needs to shower and hit the bed after a full day of sightseeing/shopping.} \comm{The rooms in the hotel are clean, modern} \cont{and air-conditioning works well.} \cont{The food served in the restaurant was varied and varied. The hotel provided a free shuttle service to Mongkok and the harbour area. The Ladies Market in Mong Kok is a pleasant walk away but the hotel bus route takes us close by.}\\
        \bottomrule\toprule
        \multicolumn{2}{l}{\textbf{\model{} w/o \method}}\hfill Intra-ROUGE1/2/L = (45.03, 11.64, 26.18)\\\midrule
        \textbf{Entity ID: 305813} & \textbf{Entity ID: 305947} \\
        ~\cont{The hotel has a great spa and sauna facilities} \comm{and is centrally located to other attractions.} \cont{It is also worth booking into the club floor for the daily cocktail hour} \comm{and internet access.} \textit{The hotel could not do enough for you.} \textit{The staff at the hotel were not very much co-operative and could not help enough.} \cont{The spa facilities were of a very high standard} \comm{and the food was of a really good quality.} \textit{The taxi drivers take advantage of the hotel's direct access to the mall and cinema next door.} & ~\cont{The Metropark Kowloon is a good hotel to stay at for a week or longer.} \comm{The hotel is in a good location} \textit{and only a short walk away from the main shopping areas of Hong Kong.} \comm{The rooms in the hotel are clean} \cont{and with air conditioning but the rooms can be quite chilly compared to the humdity outside.} \comm{The staff at the hotel were very helpful and accommodating.} \cont{The buffet breakfast was really good and varied. The food served in the restaurant was really varied and tasty.} \comm{The Sip Sip bar offered a great variety of cocktails.} \\\bottomrule
    \end{tabular}
    \caption{Contrastive summaries generated by \model{} with and w/o \method{} for an example entity pair. Distinctive (common) opinions are highlighted in \cont{blue} (\comm{magenta}), and hallucinated content is in italics.}
    \label{tab:qualitative_contrastive}
\end{table*}

\begin{table*}[t]
    \centering
    \small
    \begin{tabular}{p{15cm}}
    \toprule
        \textbf{Target Entity ID: 614392 vs Counterpart Entity ID: 1022738}\\
        The Pullman hotel is ideally situated for a city/beach vacation. \specific{Port Olympic,} the Beach \specific{and Barcelonetta} are all within walking distance. \specific{The hotel has 2 great pools, one roof top pool with bar} and \textit{one rooftop pool with a bar.} \specific{It's not cheap though. The fitness centre in the hotel is tiny but there is a fitness park about 2 minutes walking distance for 16eur/day which provide a good facility.} \\\midrule
        \textbf{Target Entity ID: 614392 vs Counterpart Entity ID: 256595}\\
        The Pullman Hotel Barcelona is a stylish hotel next to the beach \specific{with impeccable customer service.} The hotel is well situated in Barcelona, \specific{not too far from the 5 star establishment, the Arts Hotel etc. The rooms in the hotel are of a good size and nicely decorated. The room has a great balcony and sea view and the bed is incredibly comfortable. The bathroom is also really luxurious. The staff at the hotel were really attentive and really go out of their way to treat all of their guests like they are royalty.The Mini bar was expensive so avoid at all costs FYI.}\\\bottomrule
    \end{tabular}
    \caption{Contrastive summaries (Entity ID: 614392) generated by \model{} with \method{} using different entities as counterpart (Entity ID: 1022738 and 256595). The \model{} can generate completely different summaries by different conditioning. \specific{Different opinions summarized} are color-coded and \textit{hallucinated content} is in \textit{italics}}
    \label{tab:different_ent}
\end{table*}

\subsection{Analysis on \method{} Design}
Our design of \method{} uses different types of distribution aggregation methods for contrastive (Eq. (\ref{eq:codec_cont})) and common summary generation (Eq. (\ref{eq:codec_comm})).
To support those intuitive designs, we examine how the quality of generated summaries is affected when different configurations in \method{} are used for each task.
The full table is presented in the Appendix.

\noindent
\paragraph{Contrastive summary generation}
First, we tested the MoE style aggregation that is used for contrastive summary generation.
Specifically, we use addition to combine the original distribution and the ratio distribution instead of multiplication: $p_{\text{cont}}^{A\setminus B}(y_t) + \delta \left( p_{\text{cont}}^{A\setminus B}(y_t) / p_{\text{cont}}^{B\setminus A}(y_t) \right)$.

With this configuration, we observe significant degradation of summarization quality (e.g., 3.14 on R1) due to a serious distribution collapse issue in the aggregated token probability distribution.
This is mainly caused by the lack of the {\em cancellation effect} obtained by the PoE style aggregation. That is, if the probability of a token were low in the ratio distribution, it would be canceled out via the {\em multiplication} operation.

We also tested another way to highlight contrastive opinions using the common summary generation model for the ratio distribution. That is, we replace the ratio distribution in Eq. (\ref{eq:codec_cont}) with $p_{\text{cont}}^{A}(y_t) / p_{\text{comm}}^{A\cap B}(y_t)$.

This configuration shows competitive performance as the original \model{} in both self-supervised and few-shot settings, supporting the effectiveness of \method{} regardless of the specific model choice.
However, this configuration requires an additional base common opinion summarization model $p_{\text{comm}}^{A\cap B}$.
Thus, we decided to use the simpler configuration Eq. (\ref{eq:codec_cont}) as the default setting.

\noindent
\paragraph{Common summary generation}
Similarly, we verified the effectiveness of the PoE style configuration for common summary generation. That is, we use multiplication instead of addition:
$p^{A\cap B}_{\text{comm}}(y_t) \prod_{{\scriptscriptstyle E \in \{A,B\}}} p_{\text{cont}}^{E}(y_t)^\gamma$.

This configuration consistently under-performs with the original \method{} for both summarization and inter-distinctiveness scores.
This indicates that PoE focuses too much on the tokens that are likely to appear in both contrastive and common summaries, and thus it tends to generate overly generic summaries.

\subsection{Qualitative Analysis}
\paragraph{Does \method{} generate more distinctive opinions?}
Table~\ref{tab:qualitative_contrastive} shows example generations by \model{} with and w/o \method{} for contrastive summary generation. 
While both models generate summaries that are consistent with the target entity reviews, the summaries generated by \model{} w/o \method{} tend to contain \comm{common opinions} that are true for both of the entities and are against the purpose of \taskname{}.
On the contrary, \model{} actively generates opinions that can only be generated by the target entity's model $p_{\text{cont}}^{A}$, and thus the generated summary contains more \cont{contrastive opinions} for users to compare the entities.

\paragraph{Do different pairs yield different summaries?}
Distinctive opinions can change when the entity to be compared changes. 
Table~\ref{tab:different_ent} shows the generated contrastive summaries using different entities as counterpart.
As in the previous example, \model{} can generate generally consistent summaries with the target entity reviews in each setting, but also it uses \specific{different opinions} to generate summaries.
In other words, the model can highlight different opinions by comparing them with different entities, and thus generate summaries that include significantly different opinions for each.

\section{Related Work}
\noindent
{\bf Abstractive opinion summarization} aims to generate a fluent summary that reflects salient opinions in input reviews. Due to the lack of sufficient amount of reference summaries, the most common solution is the unsupervised approach~\interalia{chu2019meansum,brazinskas-etal-2020-unsupervised,amplayo-lapata-2020-unsupervised,suhara-etal-2020-opiniondigest,amplayo2021unsupervised,iso21emnlpfindings,elsahar-etal-2021-self,im-etal-2021-self,wang-wan-2021-transsum,Isonuma:2021:TACL,ke2022consist}.

Recent opinion summarization models
use the few-shot learning approach that fine-tunes a pre-trained Transformer model with a limited amount of pairs of input reviews and reference summaries. 
\citet{brazinskas-etal-2020-shot} and \citet{oved-levy-2021-pass} show that the few-shot learning approach substantially outperforms unsupervised learning models. 

All the existing methods listed above are designed for general opinion summarization and, thus, are not necessarily suitable for \taskname, as shown in the experiments.

\noindent
{\bf Comparative summarization}
There is a line of work on extracting comparative information from single/multiple documents.
\citet{lerman-mcdonald-2009-contrastive} defined the contrastive summarization problem and presented early work on the problem. Their method selects sentences so that two sets of summaries can highlight differences.
\citet{wang2013comparative} developed an extractive summarization method for a problem of Comparative Document Summarization, which is to select the most discriminative sentences from a given set of documents.
\citet{bista2019comparative} tackled a similar problem by selecting documents that represent in-cluster documents while they are useful to distinguish from other clusters.

Other studies~\cite{kim2009generating,huang-etal-2011-comparative,sipos2013generating,ren2017comparative} tackled similar tasks by developing extracting sentences/phrases from given sets of documents for comparative document analysis.
Topic models have also been used to capture comparative topics for better understanding text corpora, but they do not generate textual summaries~\cite{ren2015summarizing,he2016exploring,ibeke-etal-2017-extracting}.

Our work differs from the existing work in two points.
First, none of them focuses on generating common summaries. Second, all of the previous studies for contrastive summary generation use the extractive approach. To the best of our knowledge, we are the first to develop an opinion summarization model and a benchmark for the abstractive contrastive and common summary generation tasks.

\section{Conclusions}
In this paper, we propose a new \taskname{} task, which aims to generate contrastive and common summaries from reviews of a pair of entities, to help the user answer the question ``Which one should I pick?''
To this end, we develop a comparative summarization framework \model{}, which consists of two base summarization models; \model{} also implements \method{}, which jointly uses the token probability distribution of each model to generate more distinctive summaries in the decoding step. 

For evaluation, we created a \taskname{} benchmark \corpus{} based on the TripAdvisor review corpus. 
Experimental results on \corpus{} show that \model{} with \method{} significantly outperforms existing opinion summarization models with respect to both summarization quality and distinctiveness.
We also confirm that \method{} successfully augments \model{}, so it can generate more distinctive contrastive and common summaries than other models through comprehensive analysis. 

\section*{Acknowledgements}
We thank Wang-Chiew Tan for valuable inputs that helped create initial ideas and the anonymous reviewers for their insightful comments.

\bibliography{acl}
\bibliographystyle{acl_natbib}

\clearpage
\appendix

\section{Comparative Opinion Summarization}

Table~\ref{tab:task} shows the task comparison against existing summarization tasks. \Taskname{} is the first work that aims to generate abstractive summaries for contrastive and common opinions.

\begin{table}[t]
    \centering
    \footnotesize
    \setlength{\tabcolsep}{3pt}
    \begin{tabular}{l|ccc}
    \toprule
     & Abst. & Cont. & Comm.\\\midrule
    \citet{chu2019meansum} & \ding{51} &  &  \\
    \citet{brazinskas-etal-2020-shot,brazinskas-etal-2020-unsupervised} & \ding{51} &  & \\
    \citet{lerman-mcdonald-2009-contrastive} &  & \ding{51} &  \\
    \citet{huang-etal-2011-comparative} &  & \ding{51} &  \\
    \citet{sipos2013generating} &  & \ding{51} &  \\
    \citet{ren2017comparative}$^\dagger$ & & \ding{51} & \ding{51}\\\midrule
    {\bf This work}
    & \ding{51} & \ding{51} & \ding{51} \\\bottomrule
    \end{tabular}
    \caption{Novelty of \taskname{} against existing (opinion) summarization tasks. This work is the first task that targets to generate abstractive summaries (Abst.) for contrastive (Cont.) and common (Comm.) opinions. Note that \citet{ren2017comparative} extract keywords instead of creating textual summary.}
    \label{tab:task}
\end{table}

\section{The \corpus{} Corpus}

\subsection{Entity-Pair Selection}
For comparative opinion summarization, each of the selected entity pairs should always be comparable. To achieve this goal, we leverage the meta information of hotels in the TripAdvisor corpus to make sure that the selected entity pairs always locate in the same region (e.g., Key West of Florida).

\subsection{Annotation}
The input for each entity pair includes 16 reviews, which may be too difficult for human writers to write summaries from. Thus, we used a two-stage annotation method to ensure the quality of reference summaries.

\paragraph{Sentence Annotation}
Our first annotation task focuses on obtaining a set of sentences that contain contrastive and common opinions.
Since the average number of sentences in each entity pair (90 in \corpus) was too many to annotate at once, we grouped sentences based on their aspect category to further simplify the annotation task, 
In particular, we first split input reviews into sentences. Then, we grouped sentences into 6 aspect categories (i.e., general, staff, food, location, room, and others) using a BERT-based aspect category classifier trained with 3K labeled sentences.
By doing so, we ensure that the number of sentences annotators need to review each time is no more than 20. 
For every sentence from entity $e_A$ ($e_B$), we asked human annotators to compare it against a group of reference sentences of the same aspect category from entity $e_B$ ($e_A$) and to distinguish whether it contains any common opinions that also appear in the reference sentences.

\begin{figure*}[t]
    \centering
    \includegraphics[width=0.95\linewidth]{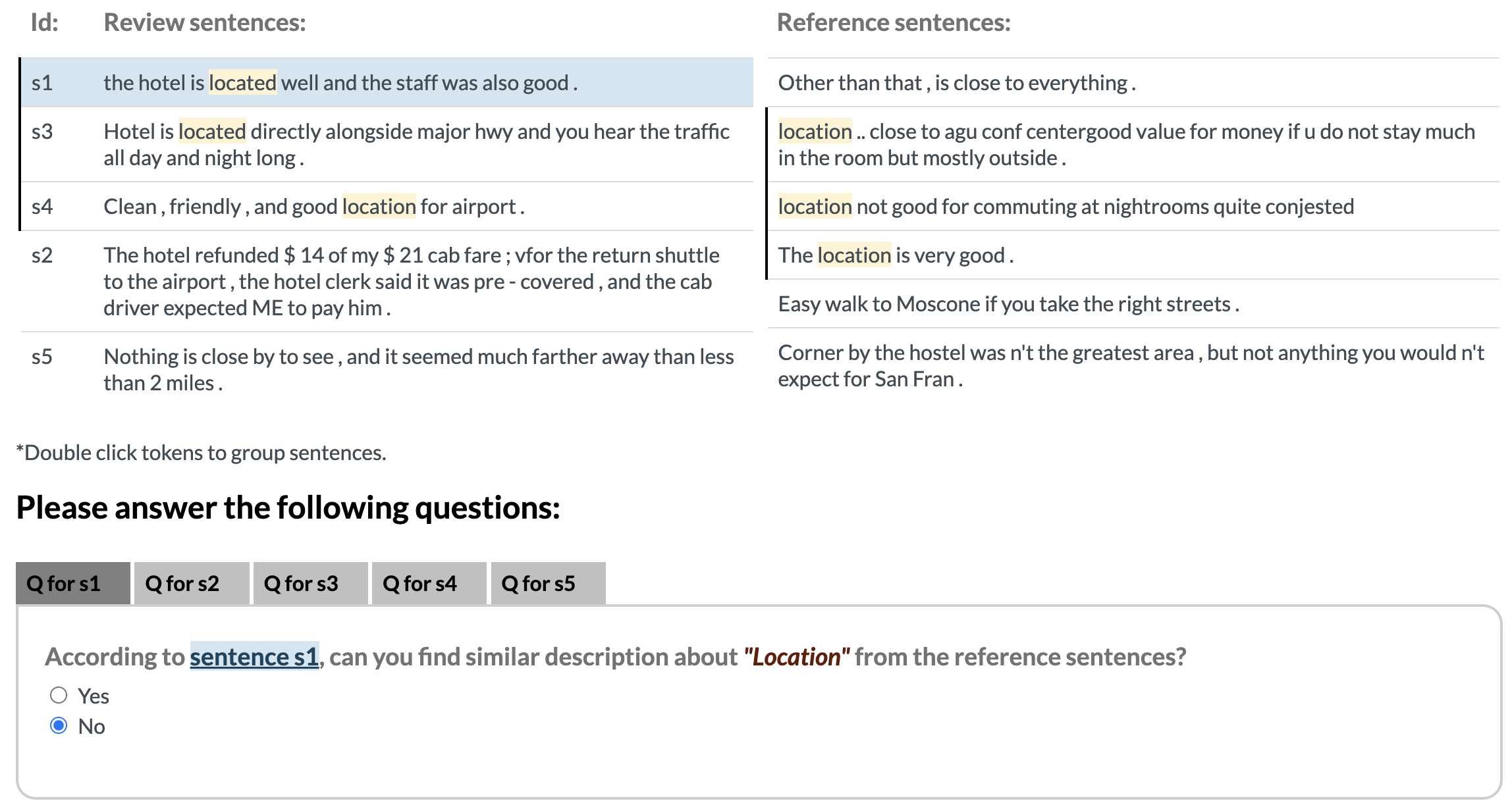}
    \caption{Sentence Annotation Task. By showing sentences of the same aspect category, it is easier for annotators to compare two group of sentences (from two entities). To further facilitate the annotation process, we also provide several additional features, such as allowing workers to group sentences that contain the same token through double clicking, and to highlight sentences through hovering over the sentence label. }
    \label{fig:sentence_annotation}
\end{figure*}

For the sentence annotation task, we hired 6 annotators from Appen's\footnote{\url{https://appen.com/}} expert worker pool with a cost of $\$0.85$ per annotation. We collected 3 annotations for each review and finalized the label through a majority vote. We obtained labels suggesting whether it contains contrastive or common opinions for every sentence in the entity pairs with the sentence annotation task. The inter annotator agreement (Fleiss' kappa) is $0.5048$. The task interface is shown in Figure~\ref{fig:sentence_annotation}.

\paragraph{Summary Collection}
In the second annotation task, we first asked human writers to write aspect-based summaries.
To exclude unreliable labels obtained in the previous step, we displayed two sets of sentences, one from each entity, to human writers for the summary collection task. This helps human writers ignore irrelevant or incorrectly labeled sentences.
For example, to obtain the contrastive summary for aspect location, we first show two corresponding sets of contrastive sentences from both $e_A$ and $e_B$ based on the labels we collected in the previous annotation step.
Then, we asked human writers to write two contrastive summaries for $e_A$ and $e_B$, respectively. Similarly, we asked human writers to write a single common summary by showing two corresponding sets of common sentences.
By doing so, we obtained aspect-based summaries for each entity pair, which are then concatenated into a reference summary. 

\begin{figure*}[t]
    \centering
    \includegraphics[width=0.95\linewidth]{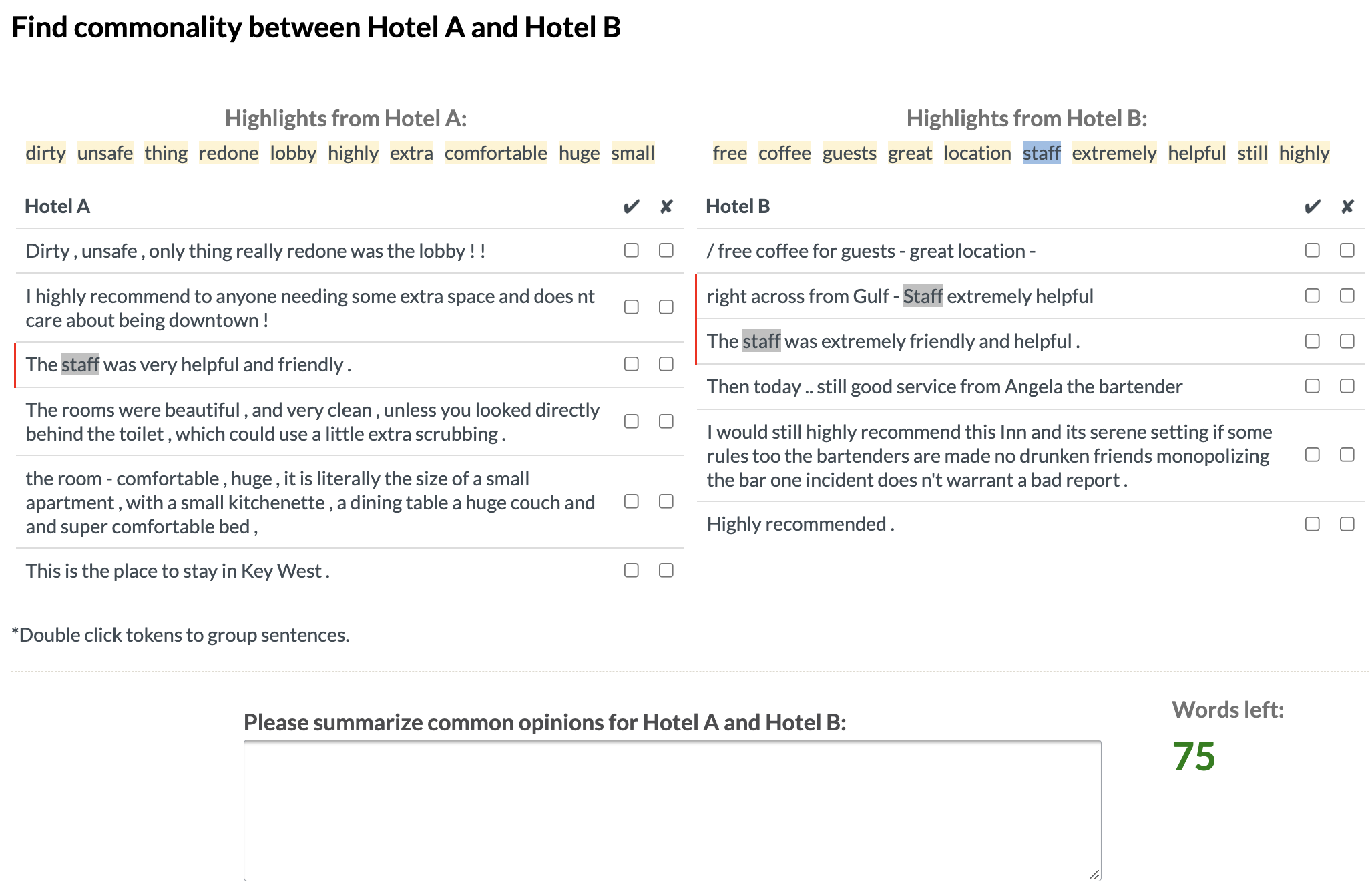}
    \caption{Summary Collection Task. We show workers two group of sentences based on labels we collected from the sentence annotation task. Similar features, such as allowing workers to group sentences that contain the same token through double clicking, are also supported in this task. }
    \label{fig:summary_collection}
\end{figure*}

Similar to the sentence annotation task, we also hired workers from Appen's expert worker pool. We hired 4 expert workers for the task with an hourly rate of $\$18$.
For contrastive (common) summaries, annotators requires in average $208$ ($107$) seconds to complete a summary. %
For every entity pair, we collected 3 reference summaries for each of {\em two} contrastive summary generation and {\em one} common summary generation tasks. The task interface is shown in Figure~\ref{fig:summary_collection}.
Since it is a text summarization task, we report their agreement via ROUGE/BERTScore in Table~\ref{tab:main} as \textit{Human upper bound}. As shown, annotators acquires $47.37$ ROUGE-1 and $37.69$ BERTScore, both are significantly higher than the baseline approaches.

\begin{table*}[t]
    \centering
    \footnotesize
    \begin{tabular}{l|cccccccc}
    \toprule
        \multirow{2}{*}{\textbf{Contrastive}} & \multicolumn{4}{c}{Summarization F1$\uparrow$} & \multicolumn{4}{c}{Intra-Distinctiveness F1$\downarrow$}\\
        & R1 & R2 & RL & BS & R1 & R2 & RL & BS \\\midrule
        \textbf{Self-supervised} \\
        \quad Original (Eq. (\ref{eq:codec_cont})) & {\bf 40.78} & {\bf 10.66} & {\bf 21.53} & \textbf{29.90} & \textbf{43.89} & \textbf{17.67} & \textbf{29.13} & \textbf{34.90} \\
        \qquad $p_{\text{cont}}^B \rightarrow p_{\text{comm}}^{A\cap B}$ & 40.60 & 10.50 & 21.36 & 29.69 & 46.95 & 18.03 & 30.00 & 39.73 \\ %
        \quad Mixture-of-Experts & 3.14 & 0.35 & 3.02 & -48.33 & 100.00 & 100.00 & 100.00 & 100.00\\
        \textbf{Few-shot} \\
        \quad Original (Eq. (\ref{eq:codec_cont})) & 42.22 & {\bf 12.11} & {\bf 24.13} & \textbf{35.63} & \textbf{35.02} & 8.39 & 21.74 & 28.23 \\
        \qquad $p_{\text{cont}}^B \rightarrow p_{\text{comm}}^{A\cap B}$ & \textbf{42.35} & 11.52 & 23.58 & 34.51 & 36.19 & \textbf{7.96} & \textbf{21.03} & \textbf{27.08} \\
        \quad Mixture-of-Experts & 3.14 & 0.35 & 3.02 & -48.33 & 100.00 & 100.00 & 100.00 & 100.00 \\\midrule\midrule
        \multirow{2}{*}{\textbf{Common}} & \multicolumn{4}{c}{Summarization F1$\uparrow$} & \multicolumn{4}{c}{Inter-Distinctiveness F1$\downarrow$}\\
         & R1 & R2 & RL & BS & R1 & R2 & RL & BS \\\midrule
         \textbf{Self-supervised} \\
         \quad Original (Eq. (\ref{eq:codec_comm})) & \textbf{41.13} & \textbf{15.25} & \textbf{30.60} & \textbf{54.65} & \textbf{50.28} & \textbf{30.12} & \textbf{44.46} & \textbf{59.81}\\
         \quad Product-of-Experts & 39.68 & 14.36 & 28.52 & 52.91 & 57.15 & 36.10 & 48.53 & 61.43\\
         \textbf{Few-shot} \\
         \quad Original (Eq. (\ref{eq:codec_comm})) & \textbf{46.80} & \textbf{20.68} & \textbf{35.62} & \textbf{61.52} & \textbf{65.14} & \textbf{43.03} & \textbf{55.15} & \textbf{70.24} \\
         \quad Product-of-Experts & 44.68 & 18.32 & 34.18 & 59.76 & 70.11 & 52.23 & 67.61 & 76.26\\
    \bottomrule
    \end{tabular}
    \caption{Summarization performance and Intra/Inter-Distinctiveness scores by \model{} with different \method{} configurations.
    }
    \label{tab:ablation}
\end{table*}

\begin{table*}[t]
    \centering
    \small
    \begin{tabular}{l|cccccccc}
    \toprule
        \multirow{2}{*}{\textbf{Contrastive}} & \multicolumn{4}{c}{Summarization F1$\uparrow$} & \multicolumn{4}{c}{Intra-Distinctiveness F1$\downarrow$}\\
        & R1 & R2 & RL & BS & R1 & R2 & RL & BS \\\midrule
        \textsc{Self-then-Few} & \textbf{43.65} & \textbf{12.83} & \textbf{24.93} & \textbf{35.42} & \textbf{39.63} & \textbf{11.80} & \textbf{25.28} & \textbf{30.72}\\
        \textsc{Multi-tasking} & 40.81 & 11.37 & 22.25 & 30.27 & 52.98 & 26.40 & 35.65 & 43.28\\
        \textsc{Only Few-shot} & 43.10 & 12.44 & 23.99 & 33.28 & 42.65 & 14.65 & 27.05 & 29.82\\\bottomrule\toprule
        \multirow{2}{*}{\textbf{Common}} & \multicolumn{4}{c}{Summarization F1$\uparrow$} & \multicolumn{4}{c}{Inter-Distinctiveness F1$\downarrow$}\\
        & R1 & R2 & RL & BS & R1 & R2 & RL & BS \\\midrule
        \textsc{Self-then-Few} & \textbf{45.90} & 19.59 & \textbf{34.40} & \textbf{59.32} & 53.87 & \textbf{29.08} & \textbf{37.94} & 60.96\\
        \textsc{Multi-tasking} & 44.64 & 17.36 & 33.87 & 58.37 & \textbf{53.13} & 29.87 & 42.45 & \textbf{59.12}\\
        \textsc{Only Few-shot} & 42.56 & \textbf{20.07} & 32.11 & 57.87 & 62.08 & 44.14 & 49.59 & 64.30\\
        \bottomrule
    \end{tabular}
    \caption{Comparisons of different few-shot learning strategies for contrastive and common opinion summarization. \textsc{Self-then-Few} further fine-tunes the self-supervised models using few-shot training data;
    \textsc{Multitasking} trains base summarization models with the pseudo review-summary data (used for self-supervised models) and few-shot training data jointly; \textsc{Only Few-shot} fine-tunes a pre-trained model (i.e., \texttt{led-base-16384} in this paper) only using few-shot training data.}
    \label{tab:fewshot_config}
\end{table*}

\section{Additional Experimental Details}

\subsection{Training details}
Major hyper-parameters for training models are reported in Table in \ref{tab:contrastive_param} and \ref{tab:common_param} following the "Show-You-Work" style suggested by \citet{dodge-etal-2019-show}.

\begin{table*}[ht]
    \centering
    \small
    \begin{tabular}{cc}
        \toprule
       \textbf{Computing infrastructure} & NVIDIA A100\\
       \midrule
       \textbf{Training duration} & Self-supervision: 12 hours, Few-shot learning: 1 hours\\
       \midrule
       \textbf{Search strategy} & Manual tuning \\\midrule
       \textbf{Model implementation} & \url{https://github.com/megagonlabs/cocosum}\\
       \midrule
       \textbf{Model checkpoint - self supervised} & \url{https://huggingface.co/megagonlabs/cocosum-cont-self}\\
       \midrule
       \textbf{Model checkpoint - few-shot} & \url{https://huggingface.co/megagonlabs/cocosum-cont-few}\\
       \bottomrule
    \end{tabular}

    \vspace{3mm}\begin{tabular}{cM{5cm}M{5cm}}
    \toprule
    \textbf{Hyperparameter} & \textbf{Search space} & \textbf{Best assignment} \\
    \midrule
    \# of training data for self-supervision & \emph{choice}[25k, 50k, 100k, 200k] & 50k\\
    \midrule
    \# of training steps for self-supervision & 50,000 & 50,000\\ 
    \midrule
    validation interval for self-supervision & 5,000 & 5,000\\
    \midrule 
    Few-shot learning strategy & \emph{choice}[\textsc{Self-then-Few}, \textsc{Multi-Tasking}, \textsc{Only Few-shot}] & \textsc{Self-then-Few}\\
    \midrule
    \# of training steps for few-shot learning & 1,000 & 1,000 \\
    \midrule
    validation interval for few-shot learning & 100 & 100\\
    \midrule
    batch size & 8 & 8\\
    \midrule
    initial checkpoint & \texttt{allenai/led-base-16384} & \texttt{allenai/led-base-16384} \\
    \midrule
    label-smoothing~\tiny{\cite{szegedy2016rethinking}} & \emph{choice}[0.0, 0.1] & 0.1 \\
    \midrule
    learning rate scheduler & linear schedule with warmup & linear schedule with warmup\\
    \midrule
    warmup steps for self-supervision & 1000 & 1000 \\
    \midrule
    warmup steps for few-shot learning & 100 & 100 \\
    \midrule
    learning rate optimizer & AdamW~\tiny{\cite{loshchilov2019decoupled}} & AdamW~\tiny{\cite{loshchilov2019decoupled}}\\
    \midrule
    AdamW $\beta_1$ & 0.9 & 0.9\\
    \midrule
    AdamW $\beta_2$ & 0.999 & 0.999\\
    \midrule
    learning rate & \emph{choice}[1e-5, 1e-4, 1e-3] & 1e-5 \\
    \midrule
    weight decay & \emph{choice}[0.0, 1e-3, 1e-2] & 1e-3 \\
    \midrule
    gradient clip & 1.0 & 1.0 \\
    \bottomrule
    \end{tabular}
    \caption{\model{} search space and the best assignments for contrastive opinion summarization on \corpus{} dataset.}
    \label{tab:contrastive_param}
\end{table*}

\begin{table*}[ht]
    \centering
    \small
    \begin{tabular}{cc}
        \toprule
       \textbf{Computing infrastructure} & NVIDIA A100\\
       \midrule
       \textbf{Training duration} & Self-supervision: 2 hours, Few-shot learning: 30 minutes\\
       \midrule
       \textbf{Search strategy} & Manual tuning \\\midrule
       \textbf{Model implementation} & \url{https://github.com/megagonlabs/cocosum}\\
       \midrule
       \textbf{Model checkpoint - self supervised} & \url{https://huggingface.co/megagonlabs/cocosum-comm-self}\\
       \midrule
       \textbf{Model checkpoint - few-shot} & \url{https://huggingface.co/megagonlabs/cocosum-comm-few}\\
       \bottomrule
    \end{tabular}

    \vspace{3mm}\begin{tabular}{cM{5cm}M{5cm}}
    \toprule
    \textbf{Hyperparameter} & \textbf{Search space} & \textbf{Best assignment} \\
    \midrule
    \# of training data for self-supervision & \emph{choice}[1k, 5k, 10k, 20k] & 5k\\
    \midrule
    \# of training steps for self-supervision & 5,000 & 5,000\\ 
    \midrule
    validation interval for self-supervision & 500 & 500\\
    \midrule
    Few-shot learning strategy & \emph{choice}[\textsc{Self-then-Few}, \textsc{Multi-Tasking}, \textsc{Only Few-shot}] & \textsc{Self-then-Few}\\
    \midrule
    \# of training steps for few-shot learning & 1000 & 1000 \\
    \midrule
    validation interval for few-shot learning & 100 & 100\\
    \midrule
    batch size & 8 & 8\\
    \midrule
    initial checkpoint & \texttt{allenai/led-base-16384} & \texttt{allenai/led-base-16384} \\
    \midrule
    label-smoothing~\tiny{\cite{szegedy2016rethinking}} & \emph{choice}[0.0, 0.1] & 0.1 \\
    \midrule
    learning rate scheduler & linear schedule with warmup & linear schedule with warmup\\
    \midrule
    warmup steps for self-supervision & 1000 & 1000 \\
    \midrule
    warmup steps for few-shot learning & 100 & 100 \\
    \midrule
    learning rate optimizer & AdamW~\tiny{\cite{loshchilov2019decoupled}} & AdamW~\tiny{\cite{loshchilov2019decoupled}}\\
    \midrule
    AdamW $\beta_1$ & 0.9 & 0.9\\
    \midrule
    AdamW $\beta_2$ & 0.999 & 0.999\\
    \midrule
    learning rate & \emph{choice}[1e-5, 1e-4, 1e-3] & 1e-5 \\
    \midrule
    weight decay & \emph{choice}[0.0, 1e-3, 1e-2] & 1e-3 \\
    \midrule
    gradient clip & 1.0 & 1.0 \\
    \bottomrule
    \end{tabular}
    \caption{\model{} search space and the best assignments for common opinion summarization on \corpus{} dataset.}
    \label{tab:common_param}
\end{table*}

\subsection{Training Dataset for Self-Supervision}\label{app:synthetic}
We collected synthetic reviews-summary pairs from the TripAdvisor review corpus for self-supervised training.
Algorithm ~\ref{alg:train} shows the review summary pair collection procedure, which is based on \citet{elsahar-etal-2021-self} with a few modifications.

\begin{algorithm*}
\small
\caption{The algorithm for building synthetic training dataset $\mathcal{C}_{\text{synthetic}}$}\label{alg:train}
\begin{algorithmic}[1]
    \Require Raw review sets $\mathcal{C}_{\text{raw}} = \{\mathcal{R}_e\}_{e \in \mathcal{E}}$, task $T \in \{\texttt{contrastive}, \texttt{common}\}$, number of input reviews $n$, number of synthetic data size $K$
    \Ensure Synthetic reviews-summary pairs $\mathcal{C}_{\text{synthetic}}$ for task $T$
    \Procedure{\textsc{BuildTrain}}{$\mathcal{C}_{\text{raw}}$, $T$, $n$}
        \State set synthetic dataset $\mathcal{C}_{\text{synthetic}} \gets \{\}$
        \ForAll{review set $\mathcal{R}_e \in \mathcal{C}_{\text{raw}}$}
            \ForAll{review $r \in \mathcal{R}_e$}
                \State $ \mathcal{R}_{e, r} := \{r_1, \dots, r_n\} = \argmax\limits_{\substack{\mathcal{R}_e' \subset \mathcal{R}_e\setminus \{r\}: |\mathcal{R}_e'| = n,\\ ^\forall r_i \in \mathcal{R}_e': 50 \leq |r_i| \leq 150}} \sum_{i\in \mathcal{R}_e'} \text{sim}(r, r_i)$
                \If{$T = \texttt{contrastive}$ \& length of $r$ is between 100 and 150}
                    \State $\mathcal{C}_{\text{synthetic}} \gets \mathcal{C}_{\text{synthetic}} \cup \{(\mathcal{R}_{e, r}, r)\}$
                \ElsIf{$T = \texttt{common}$ \& length of $r$ is between 15 and 50}
                    \State $\mathcal{C}_{\text{synthetic}} \gets \mathcal{C}_{\text{synthetic}} \cup \{(\mathcal{R}_{e, r}, r)\}$
                \EndIf
            \EndFor
        \EndFor
        \State $\mathcal{C}_{\text{synthetic}} \gets \argmax\limits_{\substack{\mathcal{C}_{\text{synthetic}}'\subset \mathcal{C}_{\text{synthetic}},\\ |\mathcal{C}_{\text{sythetic}}'| = K}} \sum\limits_{(\mathcal{R}_{e, r}', r) \in \mathcal{C}_{\text{synthetic}}'} \sum\limits_{ i\in\mathcal{R}_{e, r}'}\text{sim}(r, r_i))$
        \If{$T = \texttt{contrastive}$}
            \State \Return $\mathcal{C}_{\text{synthetic}}$
        \ElsIf{$T = \texttt{common}$}
        \State sampling counterpart entity's reviews $\mathcal{R}_{e', r'}^{\text{CP}}$
        \State $\mathcal{C}_{\text{synthetic}}' \gets \{\}$
        \ForAll{$(\mathcal{R}_{e, r}, r) \in \mathcal{C}_{\text{synthetic}}$}
            \State $(\mathcal{R}_{e', r'}^{\text{CP}}, r') \gets \argmax\limits_{\substack{(\mathcal{R}_{e', r'}, r') \in \mathcal{C}_{\text{synthetic}}\setminus \{(\mathcal{R}_{e, r}, r)\}}} \text{sim}(r, r')$
            \State $\mathcal{C}_{\text{synthetic}}' \gets \mathcal{C}_{\text{synthetic}}' \cup \{(\mathcal{R}_{e, r}, \mathcal{R}_{e', r'}^{\text{CP}}, r)\}$
        \EndFor
        \State \Return $\mathcal{C}_{\text{synthetic}}'$
        \EndIf
    \EndProcedure
\end{algorithmic}
\end{algorithm*}

\section{Additional Evaluation Results}

\subsection{Analysis on Few-shot learning design}\label{sec:fewshot_explore}
To explore the best few-shot learning design, we tested three different learning strategies, \textsc{Self-then-Few}, \textsc{Multi-Tasking}, and \textsc{Only Few-Shot}. The \textsc{Self-then-Few} strategy further fine-tunes the self-supervised summarization model by few-shot training examples. The \textsc{Multi-Tasking} strategy is to train the summarization model with self-supervised data and the few-shot data jointly. The \textsc{Only Few-Shot} only fine-times a transformer model initialized with a pre-trained checkpoint, which is the \texttt{led-base-16384} in our case.

The experimental results show in Table~\ref{tab:fewshot_config}, and we found that while the \textsc{Only-Few-Shot} configuration shows surprisingly performs well compared to the \textsc{Multi-Tasking}, the \textsc{Self-then-Few} strategy performs generally well both on contrastive and common opinion summarizations. Thus, we adapt the \textsc{Self-then-Few} to build the base summarization models for our experiments.

\begin{table*}[t]
    \centering
    \small
    \begin{tabular}{p{7cm}p{7cm}}
        \toprule
        \multicolumn{2}{l}{\textbf{\model}}\\\midrule
        \textbf{Entity ID: 482693 \& 1547281} & \textbf{Entity ID: 202988 \& 233491} \\
        ~\comm{The staff at the hotel were very helpful and friendly.} \specific{The hotel is in a great location and close to the canal.} & ~\comm{The staff at the hotel were very friendly and helpful.} \specific{The hotel is ideally located for a stay in Florence.} \\\bottomrule\toprule
        \multicolumn{2}{l}{\textbf{\model w/o \method}}\\\midrule
        \textbf{Entity ID: 482693 \& 1547281} & \textbf{Entity ID: 202988 \& 233491} \\
        ~\comm{The staff at the hotel are very friendly and the hotel is recommended.} & ~\comm{This hotel is in an excellent location and the staff are very friendly and helpful.}\\
        \bottomrule
    \end{tabular}
    \caption{Common summaries generated by \model{} with and w/o \method{} for two example entity pairs. Entity-pair specific (common) opinions are highlighted in \specific{green} (\comm{magenta}).}
    \label{tab:qualitative_common}
\end{table*}

\end{document}